\documentclass[10pt,twocolumn,letterpaper]{article}

\usepackage{iccv}
\usepackage{times}
\usepackage{epsfig}
\usepackage{graphicx}
\usepackage{amsmath}
\usepackage{amssymb}
\usepackage{bbold}
\usepackage{amsthm}
\usepackage{multirow}
\usepackage{float}
\usepackage[table,xcdraw]{xcolor}
\usepackage{rotating}
\usepackage{cuted}
\usepackage{capt-of}

\usepackage[pagebackref=true,breaklinks=true,letterpaper=true,colorlinks,bookmarks=false]{hyperref}

\newtheorem{prop}{Proposition}
\newtheorem{definition}{Definition}

 \iccvfinalcopy % *** Uncomment this line for the final submission

\def\iccvPaperID{3923} % *** Enter the ICCV Paper ID here

\ificcvfinal\fi
\begin{document}

\newcommand{\norm}[1]{\left\lVert#1\right\rVert}

%%%%%%%%% TITLE
\title{3D Point Cloud Generative Adversarial Network Based on 
\\
Tree Structured Graph Convolutions}

\author{Dong Wook Shu\thanks{Authors contributed equally}, Sung Woo Park\footnotemark[1], and Junseok Kwon\\
School of Computer Science and
Engineering,
Chung-Ang University,
Seoul, Korea\\
%cau.ac.kr\\
{\tt\small jskwon@cau.ac.kr}
% For a paper whose authors are all at the same institution,
% omit the following lines up until the closing ``}''.
% Additional authors and addresses can be added with ``\and'',
% just like the second author.
% To save space, use either the email address or home page, not both
%\and
%Junseok Kwon\\
%Chung-Ang University\\
%cau.ac.kr\\
%{\tt\small jskwon@cau.ac.kr}
}

\maketitle

\begin{strip}
\centering
\includegraphics[width=\textwidth]{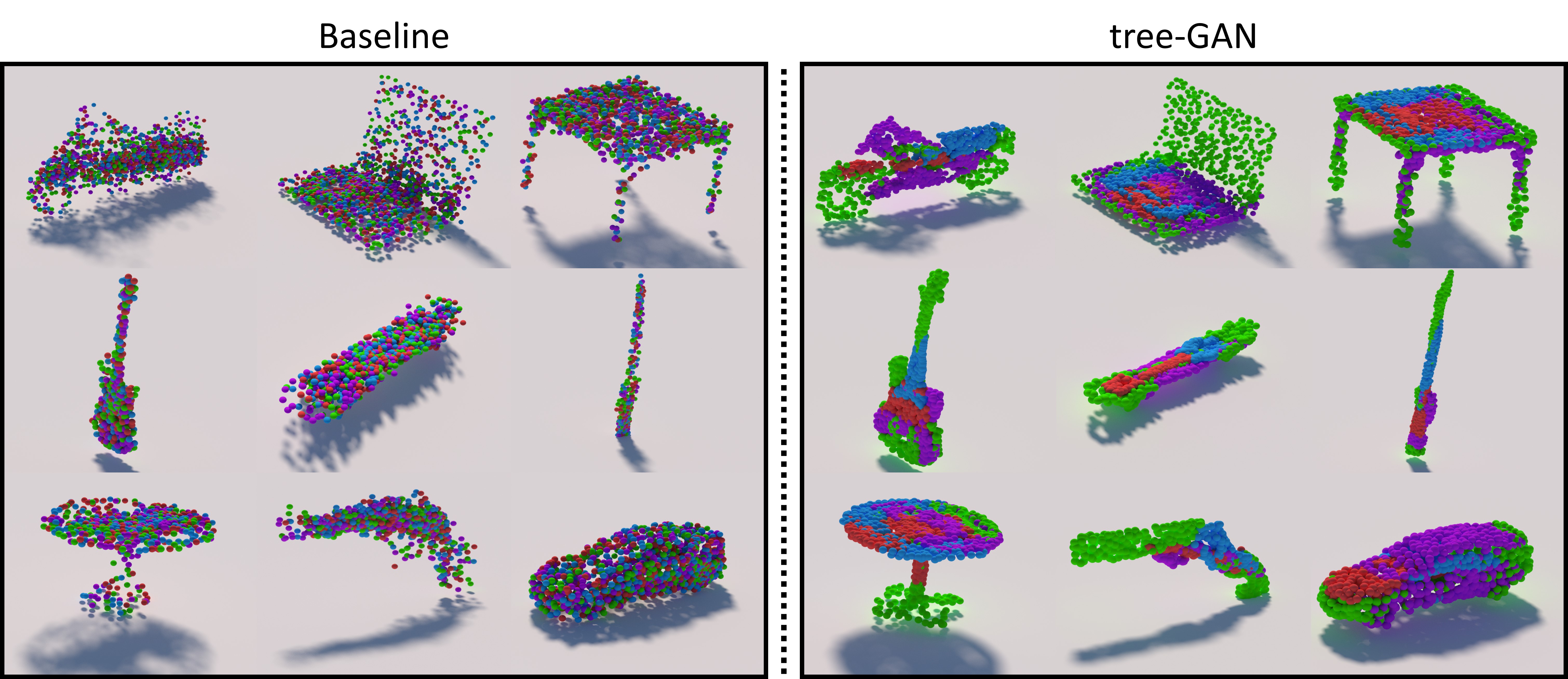}
\captionof{figure}{ \textbf{Unsupervised 3D point clouds generated by our tree-GAN for multiple classes} (\emph{e.g.}, Motorbike, Laptop, Table, Guitar, Skateboard, Knife, Table, Pistol, and Car from top-left to bottom-right).	
Our tree-GAN can generate more accurate point clouds than baseline (\emph{i.e.}, r-GAN~\cite{r-gan}), and can also produce point clouds for semantic parts of objects, which are denoted by different colors.}
\label{fig:teaser}
\end{strip}

%%%%%%%%% ABSTRACT
\begin{abstract}
In this paper, we propose a novel generative adversarial network (GAN) for 3D point clouds generation, which is called tree-GAN. To achieve state-of-the-art performance for multi-class 3D point cloud generation, a tree-structured graph convolution network (TreeGCN) is introduced as a generator for tree-GAN.
Because TreeGCN performs graph convolutions within a tree, it can use ancestor information to boost the representation power for features.  
To evaluate GANs for 3D point clouds accurately, we develop a novel evaluation metric called Fr\'echet point cloud distance (FPD).
Experimental results demonstrate that the proposed tree-GAN outperforms state-of-the-art GANs in terms of both conventional metrics and FPD, and can generate point clouds for different semantic parts without prior knowledge. 
\end{abstract}
%%%%%%%%% BODY TEXT
%%%%%%%%%%%%%%%%%%%%%%%%%%%%%%%%%%%%%%%%%%%%%%%%%%%%%%%%%%%%%%%%%%%%%%%%%%%%%%%%%%%%%%%%%%
\vspace{-5mm}
\section{Introduction}
Recently, 3D data generation problems based on deep neural networks have attracted significant research interest and have been addressed through various approaches, including image-to-point cloud~\cite{SI2PC,Jiang_GAL}, image-to-voxel~\cite{Wu_NIPS16}, image-to-mesh~\cite{Pixel2Mesh}, point cloud-to-voxel~\cite{Dai17,Zhou2018}, and point cloud-to-point cloud~\cite{FoldingNet}.
The generated 3D data has been used to achieve outstanding performance in a wide range of computer vision applications (\emph{e.g.}, segmentation~\cite{Pointnet++,Christoph14,dgcnn}, volumetric shape representation~\cite{Wu2015}, object detection~\cite{Chen15,Song16}, feature extraction~\cite{Li16}, contour detection~\cite{Hackel16}, classification~\cite{Qi17,Socher12}, and scene understanding~\cite{Kim13,Wang13}).

However, little effort has been devoted to the development of generative adversarial networks (GANs) that can generate 3D point clouds in an unsupervised manner. 
To the best of our knowledge, the only works on GANs for transforming random latent codes (\emph{i.e.}, $z$ vectors) into 3D point clouds are \cite{r-gan} and \cite{Localized-GCN}.
The method in \cite{r-gan} generates point clouds using only fully connected layers. 
The method in \cite{Localized-GCN} exploits local topology by using $k$-nearest neighbor techniques to produce geometrically accurate point clouds.
However, it suffers from high computational complexity as the number of dynamic graph updates increases. 
Additionally, it can only generate a limited number of object categories (\emph{e.g.}, chair, airplane, and sofa) using point clouds.     

In this paper, we present a novel method called tree-GAN that can generate 3D point clouds from random latent codes in an unsupervised manner. It can also generate multi-class 3D point clouds without training on each class separately (\emph{e.g.}, \cite{Localized-GCN}). 
To achieve state-of-the-art performance in terms of both accuracy and computational efficiency, we propose a novel tree-structured graph convolution network (TreeGCN) as a generator for tree-GAN.
The proposed TreeGCN preserves the ancestor information of each point and utilizes this information to extract new points via graph convolutions. A branching process and loop term with $K$ supports in TreeGCN further enhance the representation power of points. These two properties enable TreeGCN to produce more accurate point clouds and express more diverse object categories.          
Additionally, we demonstrate that using the ancestors of features in TreeGCN is more efficient computationally than using the neighbors of features in traditional GCNs.. 
Fig.\ref{fig:teaser} shows the effectiveness of our tree-GAN. 

\begin{figure*}[t]
  \centering
  \includegraphics[width=\linewidth]{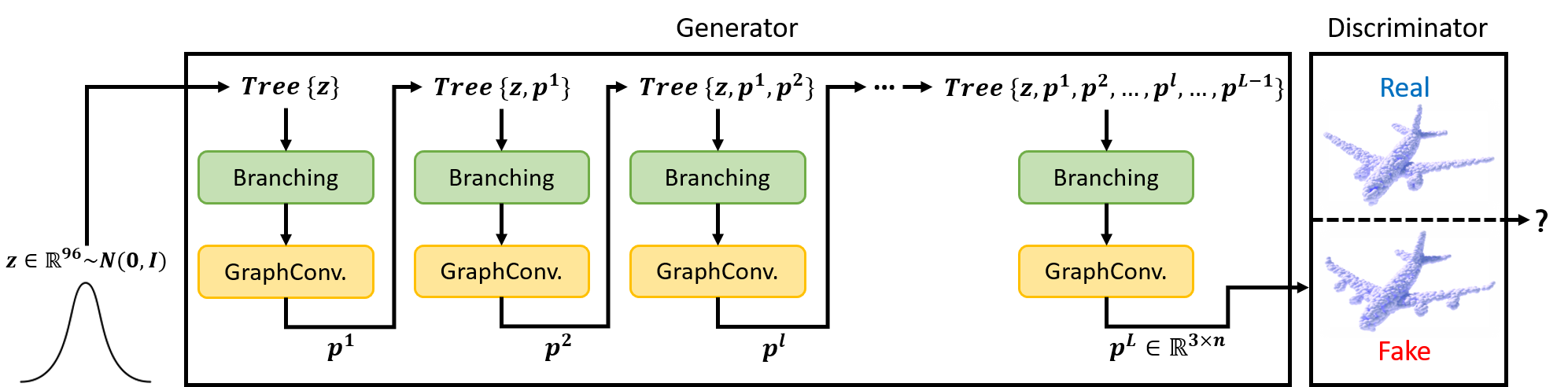}
  \caption{\textbf{Pipeline of the tree-GAN.} Our tree-GAN contains two networks, namely, discriminator (Section \ref{sec:TREE-GAN}) and generator (Section \ref{sec:TreeGCN}).  
  The generator takes a single point from a Gaussian distribution, $z \in R^{96}$, as an input. At each layer of the generator, GraphConv (Section \ref{sec:graphConv}) and Branching (Section \ref{sec:branch}) operations are performed to generate the $l$-th set of points, $p^l$. All points generated by previous layers are stored and appended to the tree of the current layer. 
  The tree begins from the root node $z$, splits into child nodes via Branching operations, and modifies nodes by GraphConv operations.  
  The generator produces 3D point clouds $x'=p^L \in R^{3 \times n}$ as outputs, where $p^L$ is the set of points at the final layer $L$ and $n$ is the total number of points.
  The discriminator differentiates between real and generated point clouds to force the generator to produce more realistic points. We use a discriminator similar to that in the r-GAN~\cite{r-gan}. 
  Please refer to supplementary materials for detailed network architectures. }
  \label{fig:pipeline}
\end{figure*}

The main contributions of this paper are fourfold.

%\begin{itemize}
\noindent $\bullet$ We present the novel tree-GAN method, which is a deep generative model that can generate multi-class 3D point clouds in unsupervised settings (Section \ref{sec:TREE-GAN}).

\noindent $\bullet$ We introduce the TreeGCN based generator. The performance of traditional GCNs can be improved significantly by adopting the proposed tree structures for graph convolutions. 
Based on the proposed tree structures, tree-GAN can generate parts of objects by selecting particular ancestors (Section \ref{sec:TreeGCN}).

\noindent $\bullet$ We mathematically interpret the TreeGCN and highlight its desirable properties (Section \ref{sec:math}). 

\noindent $\bullet$ We present the Fr\'echet point cloud distance (FPD) metric to evaluate GANs for 3D point clouds. FPD can be considered as a nontrivial extension of Fr\'echet inception distance (FID)~\cite{fid}, which has been widely used for the evaluation of GANs (Section \ref{sec:FPD}).  
%\end{itemize}

%%%%%%%%%%%%%%%%%%%%%%%%%%%%%%%%%%%%%%%%%%%%%%%%%%%%%%%%%%%%%%%%%%%%%%%%%%%%%%%%%%%%%%%%%%
%%%%%%%%%%%%%%%%%%%%%%%%%%%%%%%%%%%%%%%%%%%%%%%%%%%%%%%%%%%%%%%%%%%%%%%%%%%%%%%%%%%%%%%%%%
\section{Related Work}

%%%%%%%%%%%%%%%%%%%%%%%%%%%%%%%%%%%%%%%%%%%%%%%%%%%%%%%%%%%%%%%%%%%%%%%%%%%%%%%%%%%%%%%%%%
\noindent \textbf{Graph Convolutional Networks: }
Over the past few years, a number of works have focused on the generalization of deep neural networks for graph problems~\cite{Bruna14,Duvenaud15,Henaff15,Li_ICLR16}.
Defferrard \emph{et al.}~\cite{ChevyNet} proposed fast-learning convolutional filters for graph classification problems.
Using these filters, they significantly accelerated the spectral decomposition process, which was one of the main computational bottlenecks in traditional graph convolution problmes with large datasets. 
Kipf and Welling~\cite{firstorder-ChevyNet} introduced scalable GCNs based on first-order approximations of spectral graph convolutions for semi-supervised classification, in which convolution filters only use the information from neighboring vertices instead of the information from the entire network.

Because the aforementioned GCNs were originally designed for classification problems, the connectivity of graphs was assumed to be given as prior knowledge. 
However, this setting is not appropriate for problems of dynamic model generation. 
For example, in unsupervised settings for 3D point cloud generation, the typologies of 3D point clouds are non-deterministic. 
Even for the same class (\emph{e.g.}, chairs), 3D point clouds can be represented by various typologies. 
To represent the diverse typologies of 3D point clouds, our TreeGCN utilizes no prior knowledge regarding object models.  

%%%%%%%%%%%%%%%%%%%%%%%%%%%%%%%%%%%%%%%%%%%%%%%%%%%%%%%%%%%%%%%%%%%%%%%%%%%%%%%%%%%%%%%%%%
\noindent \textbf{GANs for 3D Point Clouds Generation: }
GANs~\cite{GAN} for 2D image generation tasks have been widely studied with great success~\cite{SeGAN, Pix2Pix,Ledig2017,Jianxin2018,Reed2016,Song2018,Wang2016,A-Fast-RCNN,Jiahui2018,Zhang2018}, but GANs for 3D point cloud generation have rarely been studied in the computer vision field.
Recently, Achlioptas \emph{et al.}~\cite{r-gan} proposed a GAN for 3D point clouds called r-GAN, generator of which is based on fully connected layers. As fully connected layers cannot maintain structural information, the r-GAN has difficulty in generating realistic shapes with diversity.
%They pre-trained auto-encoders using conventional metrics for 3D point clouds, such as %the earth mover's distance, then fitted learned latent representations to Gaussian %mixture models.
%However, the results were highly dependent on the performance of the pre-trained %auto-encoders.
% <- should be replaced!!
Valsesia \emph{et al.}~\cite{Localized-GCN} used graph convolutions for generators for GANs.
At each layer of graph convolutions during training, adjacency matrices were dynamically constructed using the feature vectors from each vertex. 
Unlike traditional graph convolutions, the connectivity of a graph was not assumed to be given as prior knowledge.
However, to extract the connectivity of a graph, computing the adjacency matrix at a single layer incurs quadratic computational complexity $O(V^2)$ where $V$ indicates the number of vertices.
Therefore, this approach is intractable for multi-batch and multi-layer networks.

Similar to the method in \cite{Localized-GCN}, our tree-GAN requires no prior knowledge regarding the connectivity of a graph. However, unlike the method in \cite{Localized-GCN}, the tree-GAN is computationally efficient because it does not construct adjacency matrices. Instead, the tree-GAN uses ancestor information from the tree to exploit the connectivity of a graph, in which only a list of tree structure is needed. 

%%%%%%%%%%%%%%%%%%%%%%%%%%%%%%%%%%%%%%%%%%%%%%%%%%%%%%%%%%%%%%%%%%%%%%%%%%%%%%%%%%%%%%%%%%
\noindent \textbf{Tree-structured Deep Networks: }
There have been several attempts to represent convolutional neural networks or long short-term memory using tree structures~\cite{Cheng18,kim_ICML17,Murdock16,Nam16,Roy18}.
However, to the best of our knowledge, no previous methods have used tree structures for either graph convolutions or GANs. 
For example, Gadelha \emph{et al.}~\cite{multi-res-vae} used tree-structured networks to generate 3D point clouds via variational autoencoder (VAE). 
However, this method needed the assumption that inputs are the 1D-ordered lists of points obtained by space-partitioning algorithms such as K-dimensional tree and random projection tree~\cite{rp}.
Thus, it required additional preprocessing steps for valid implementations. 
Because its network only comprised 1D convolution layers, the method could not extract the meaningful information from unordered 3D point clouds. 
In contrast, the proposed tree-GAN can not only deal with unordered points, but also  extract semantic parts of objects.

%%%%%%%%%%%%%%%%%%%%%%%%%%%%%%%%%%%%%%%%%%%%%%%%%%%%%%%%%%%%%%%%%%%%%%%%%%%%%%%%%%%%%%%%%%
%%%%%%%%%%%%%%%%%%%%%%%%%%%%%%%%%%%%%%%%%%%%%%%%%%%%%%%%%%%%%%%%%%%%%%%%%%%%%%%%%%%%%%%%%%
\section{3D Point Cloud GAN} \label{sec:TREE-GAN}
Fig.\ref{fig:pipeline} presents the pipeline of the proposed tree-GAN. 
To generate 3D point clouds $x'$ from latent code $z$, we utilize the objective function introduced in Wasserstein GAN~\cite{WGAN}. 
The loss function of a generator, $L_{gen}$, is defined as  
\begin{equation} \label{eq_loss_gen}
    L_{gen} = -\mathbb{E}_{z\sim \mathcal{Z}}[D(G(z))],
\end{equation}
where $G$ and $D$ denote the generator and discriminator, respectively, and $\mathcal{Z}$ represents a latent code distribution.
We design $\mathcal{Z}$ with a Normal distribution, $z \in \mathcal{N}(\mathbf{0}, I)$. 
The loss function of a discriminator, $L_{disc}$, is defined as  
\begin{equation} \label{eq_loss_disc}
\begin{split}
    L_{disc} &= \mathbb{E}_{z\sim \mathcal{Z}}[D(G(z))] - \mathbb{E}_{x\sim \mathcal{R}}[D(x)] 
    \\
    &+ \lambda_{gp} \mathbb{E}_{\hat{x}}[ (\norm{ \nabla_{\hat{x}} D(\hat{x})}_2 - 1)^2 ],
\end{split}
\end{equation}
where $\hat{x}$ are sampled from line segments between real and fake point clouds, $x'\sim G(z)$ and $x$ denote generated and real point clouds, respectively, and $\mathcal{R}$ represents a real data distribution.
In \eqref{eq_loss_disc}, we use a gradient penalty to satisfy the $1$-Lipschitz condition~\cite{WGAN-GP}, where $\lambda_{gp}$ is a weighting parameter.

%%%%%%%%%%%%%%%%%%%%%%%%%%%%%%%%%%%%%%%%%%%%%%%%%%%%%%%%%%%%%%%%%%%%%%%%%%%%%%%%%%%%%%%%%%
%%%%%%%%%%%%%%%%%%%%%%%%%%%%%%%%%%%%%%%%%%%%%%%%%%%%%%%%%%%%%%%%%%%%%%%%%%%%%%%%%%%%%%%%%%
\section{Proposed TreeGCN} \label{sec:TreeGCN}
To implement $G$ in \eqref{eq_loss_gen}, 
we consider multi-layer graph convolutions with first-order approximations of the Chebyshev expansion introduced by~\cite{firstorder-ChevyNet} as follows:
\begin{equation} \label{eq:gcn}
    p_i^{l+1} = \sigma\left( W^{l} p_i^l + \sum_{q_j^l \in N(p_i^l)}U^{l} q_j^l + b^l\right),
\end{equation}
where $\sigma(\cdot)$ is the activation unit, $p_i^l$ is the $i$-th node in the graph (\emph{i.e.}, 3D coordinate of a point cloud) at the $l$-th layer, $q_j^l$ is the $j$-th neighbor of $p_i^l$, and $N(p_i^l)$ is the set of all neighbors of $p_i^l$.
Then, $z$ and $x'$ in \eqref{eq_loss_disc} can be represented by $[p_1^0]$ and $[p_1^L~p_2^L~\cdots~p_n^L]$, respectively, where $L$ is the final layer and $n$ is the number of points at $L$.

During training, GCNs find the best weights $W^{l}$ and $U^{l}$ and best bias $b^l $ at each layer, then generate 3D coordinates for point clouds by using these parameters to ensure similarity to real point clouds.  
The first and second terms in \eqref{eq:gcn} are called the \textit{loop} and \textit{neighbors} terms, respectively.

To enhance a conventional GCN such as that used in \cite{firstorder-ChevyNet}, we propose a novel GCN augmented with tree structures (\emph{i.e.}, TreeGCN). 
The proposed TreeGCN introduces a tree structure for hierarchical GCNs by passing information from ancestors to descendants of vertices. 
The main unique characteristic of the TreeGCN is that each vertex updates its value by referring to the values of its ancestors in the tree instead of thosed of its neighbors. 
Traditional GCNs, such as those defined in~\eqref{eq:gcn}, can be considered as methods that only refer to neighbors at a single depth.  
Then, the proposed graph convolution is defined as 
\begin{equation} \label{eq:treegcn}
    p_i^{l+1} = \sigma\left(\mathbf{F}^l_K(p_{i}^l) + \sum_{q_j \in A(p_i^l)}U_j^l q_j + b^l\right), 
\end{equation}
where there are two major differences compared to \eqref{eq:gcn}. 
One is an improved conventional loop term using a subnetwork $\mathbf{F}_K^l$, where $S_i^{l+1}$ is generated by $K$ supports from $\mathbf{F}_K^l$.
We call this term a \textit{loop with $K$-supports}, as explained in Section \ref{sec:graphConv}.
The other difference is the consideration of values from, all ancestors in the tree to update the value of a current point, where $A(p_i^l)$ denotes the set of all ancestors of $p_i^l$.  
We call this term \textit{ancestors}, as explained in Section \ref{sec:graphConv}.

%%%%%%%%%%%%%%%%%%%%%%%%%%%%%%%%%%%%%%%%%%%%%%%%%%%%%%%%%%%%%%%%%%%%%%%%%%%%%%%%%%%%%%%%%%
\subsection{Advanced Graph Convolution} \label{sec:graphConv}

\begin{figure}[t]
  \includegraphics[width=\linewidth]{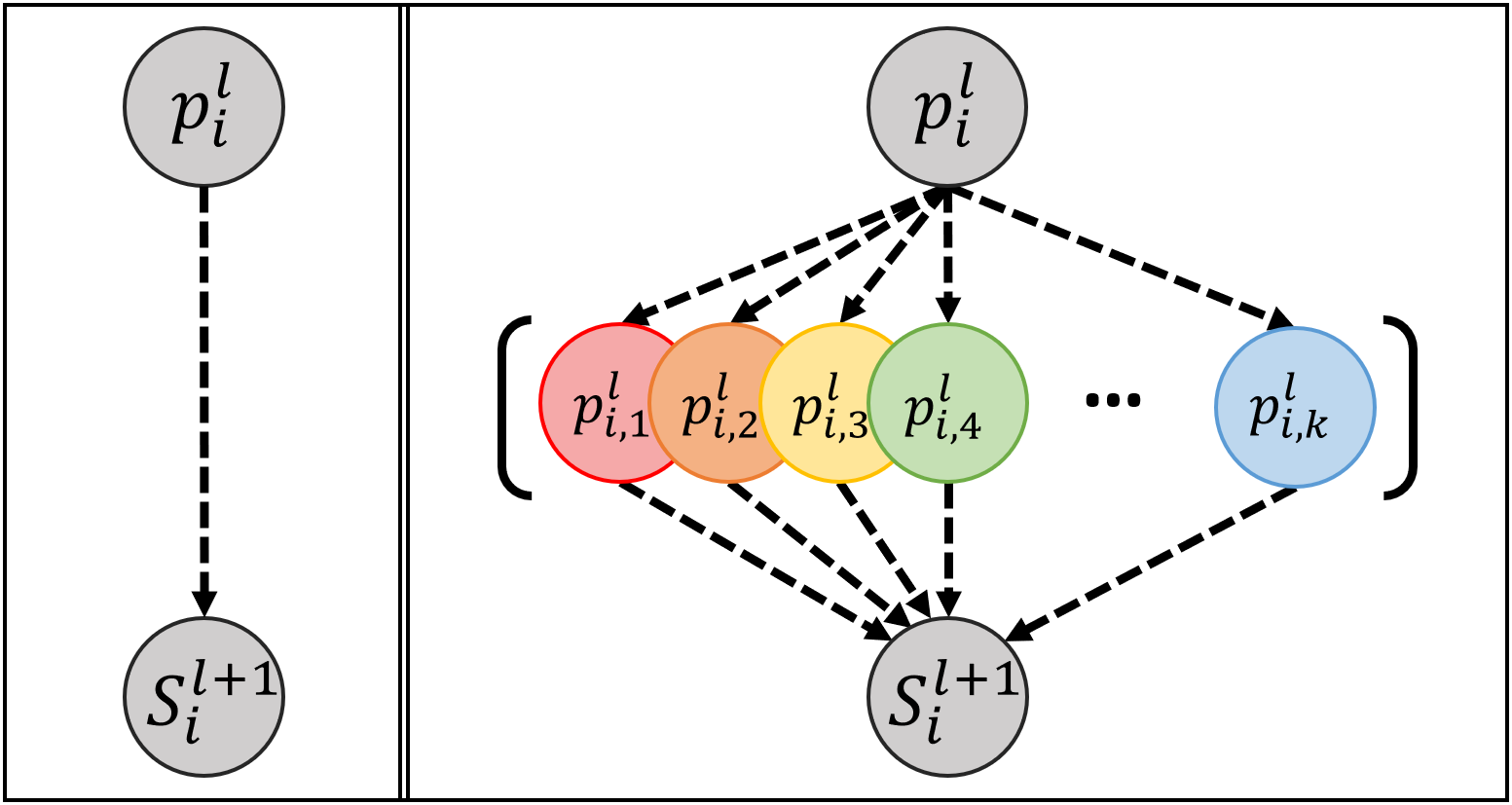}
  \caption{\textbf{Loop term with $K$-supports.} Left: a conventional loop term uses a single parameter $W^{l}$ in \eqref{eq:gcn} to learn the mapping from $p_i^l$ to $S_i^{l+1}$. Right: our loop term introduces a fully connected layer with $K$ nodes (\emph{i.e.}, $K$ supports, $p_{i,1}^l,\cdots,p_{i,k}^l$) to learn a more complex mapping from $p_i^l$ to $S_i^{l+1}$. }
  \label{fig:support}
\end{figure}

\begin{figure}[t]
  \includegraphics[width=\linewidth]{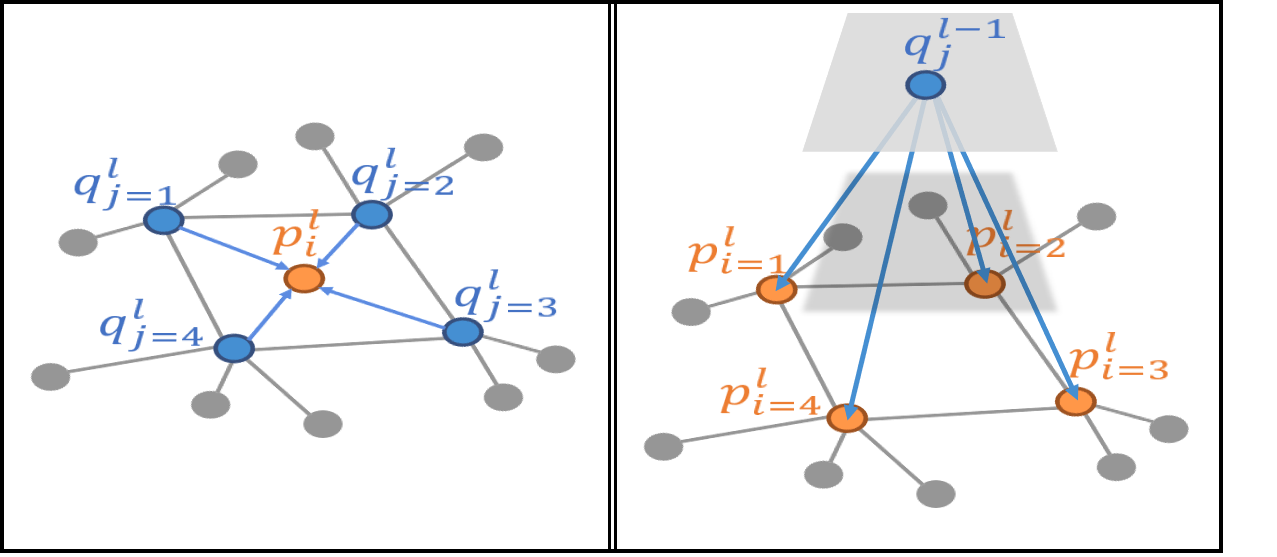}
  \caption{\textbf{Ancestor term}. Left: a conventional neighbor term uses neighbors of $p_i^l$ (\emph{e.g.}, $q_1^l,q_2^l,\cdots$) to generate $p_i^{l+1}$. Right: the proposed ancestor term uses ancestors of $p_i^l$ (\emph{e.g.}, $q_1^{l-1},q_2^{l-2},\cdots$) to generate $p_i^{l+1}$.} 
  \label{fig:ancestor}
\end{figure}

The proposed tree-structured graph convolution (\emph{i.e.}, GraphConv in Fig.\ref{fig:pipeline}) aims to modify the coordinates of points using the loop with $K$-supports and ancestor terms. 

\noindent \textbf{Loop term with $K$-supports: }
The goal of the new loop term in \eqref{eq:treegcn} is to propose the next point based on $K$ supports instead of using only the single parameter $W^l$ in \eqref{eq:gcn} as follows: 
\begin{equation} \label{eq:loop}
    S_{i}^{l+1} = \mathbf{F}^l_K(p_i^l), 
\end{equation}
where $\mathbf{F}^l_K$ is a fully connected layer containing $K$ nodes.
A conventional GCN using first-order approximations adopts a single parameter in its loop term to generate the next point from the current point. 
However, for large graphs, the representation capacity of a single parameter is insufficient for describing a complex point distribution. 
Therefore, our loop term utilizes $K$ supports to represent a more complex distribution of points, as illustrated in Fig.\ref{fig:support}.

%%%%%%%%%%%%%%%%%%%%%%%%%%%%%%%%%%%%%%%%%%%%%%%%%%%%%%%%%%%%%%%%%%%%%%%%%%%%%%%%%%%%%%%%%%
\noindent \textbf{Ancestor term: } 
For graph convolution, knowing the connectivity of a graph is very important because this information allows a GCN to propagate useful information from a vertex to other connected vertices.
However, in our point cloud generation setting, it is impossible to use prior knowledge regarding connectivity because we must be able to generate diverse typologies of point clouds, even for the same object category.  
Therefore, the dynamic 3D point generation problem cannot be addressed using traditional GCNs because such networks assume that the connectivity of a graph is given. 
As a replacement for the neighbor term in \eqref{eq:gcn}, we define the \textit{ancestor} term in \eqref{eq:treegcn} as follows:
\begin{equation} \label{eq:ancestor}
    \sum_{q_j \in A(p_i^l)}U_j^l q_j, 
\end{equation}
where $A(p_i^l)$ denotes the set of all ancestors of $p_i^l$. 
This term combines all information from the ancestors $q_j$ through a linear mapping $U_j^l$. 
Because each ancestor belongs to a different feature space at a different layer, our ancestor term can fuse all information from previous layers and different feature spaces. 
To generate the next point, the current point refers to its ancestors in various feature spaces to find the best mapping $U_j^l$ to combine ancestor information effectively. 
By using this new \textit{ancestor} term, our tree-GAN obtains several desirable mathematical properties, as explained in Section \ref{sec:math}.
Fig.\ref{fig:ancestor} illustrates the graph convolution process with the ancestor term.

%%%%%%%%%%%%%%%%%%%%%%%%%%%%%%%%%%%%%%%%%%%%%%%%%%%%%%%%%%%%%%%%%%%%%%%%%%%%%%%%%%%%%%%%%%
\subsection{Branching} \label{sec:branch}

\begin{figure}[t]
  \includegraphics[width=\linewidth]{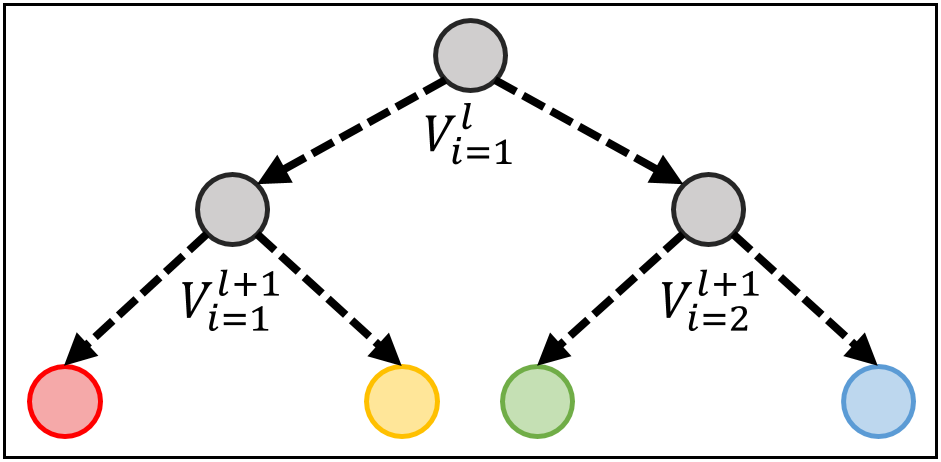}
  \caption{\textbf{Example of branching with degree 2.}}
  \label{fig:branch}
\end{figure}

Branching is a procedure for increasing the total number of points and is similar to up-sampling in 2D convolution.
In branching, $V_i^{l+1}$ transforms a single point $p_i^l \in R^3$ into $d_l$ child points, where $[V_i^{l+1} \cdot p_i^l] \in R^{3 \times d_l}$. Therefore,
\begin{equation} \label{eq:branch}
    p_j^{l+1} = [V_i^{l+1} \cdot p_i^l]_j,~\text{for}~j=1,\cdots,d_l 
\end{equation}
where $[A]_j$ denotes the $j$-th column of matrix $A$.
Then, the total number of points in the $(l+1)$-th layer is $|p^l| \times d_l$, where $|p^l|$ is the number of points in the $l$-th layer.
In our experiments, we use different branching degrees for different layers (\emph{e.g.}, $\{d_l\}_{l=1}^{7} = \{1, 2, 2, 2, 2, 2, 64\}$).
Note that the number of points in the final layer is $\prod_{l=1}^{7} d_l = 2048$. 
Fig.\ref{fig:branch} presents an example of branching with degree $2$.  

%%%%%%%%%%%%%%%%%%%%%%%%%%%%%%%%%%%%%%%%%%%%%%%%%%%%%%%%%%%%%%%%%%%%%%%%%%%%%%%%%%%%%%%%%%
%%%%%%%%%%%%%%%%%%%%%%%%%%%%%%%%%%%%%%%%%%%%%%%%%%%%%%%%%%%%%%%%%%%%%%%%%%%%%%%%%%%%%%%%%%
\section{Mathematical Properties} \label{sec:math}
In this section, we mathematically analyze the geometric relationships between generated points and demonstrate how these relationships are formulated in the output Euclidean space via tree-structured graph convolutions. 

\begin{prop} \label{prop:1}
Let $p_s^L$ and $p_d^L$ in \eqref{eq:treegcn} be generated points that share the same parents and different parents, respectively, with $p_i^L$ in the final layer $L$.
Let $S_{i}^l$ in \eqref{eq:loop} be the loop of point $p_i^l$ at the $l$-th layer. 
Hereafter, we omit the superscripts of $p_s^l$ and $S_{i}^l$ if the superscript $l$ indicates the fianl layer $L$.
Then, 
\begin{equation} \label{eq:d_1}
    \norm{p_s - p_i}^2 = \norm{S_{s} - S_{i}}^2,
\end{equation} 
and 
\begin{equation} \label{eq:d_2}
    \norm{p_d - p_i}^2 \leq \sum_{l=1}^{L-1}  \norm{S_{A_l(p_d)}^l - S_{A_l(p_i)}^l}^2 + \norm{S_{d} - S_{i}}^2,
\end{equation}
where $A_l(p)$ are all ancestors of point $p$ in the $l$-th layer.
For simplicity, we ignore the \textit{branch} process and $U_j^l$ in \eqref{eq:loop}. 
\end{prop}
Based on Proposition~\ref{prop:1}, we can prove that following two statements are true:

\noindent $\bullet$ \textbf{The geometric distance between two points is determined by the number of shared ancestors.}
If two points $p_d$ and $p_i$ have different ancestors, then the geometric distance between these points is calculated as the sum of differences between their ancestors in each layer $l$  $\left(\emph{i.e.},~\norm{S_{A_l(p_d)}^l - S_{A_l(p_i)}^l}^2~\text{in}~\eqref{eq:d_2}\right)$ and the differences between their loops $\left(\emph{i.e.},~\norm{S_{d} - S_{i}}^2~\text{in}~\eqref{eq:d_2}\right)$. 
Thus, as their ancestors become increasingly different, the geometric distance between two points increases. 

\noindent $\bullet$ \textbf{Geometrically related points share the same ancestors.} 
If two points $p_s$ and $p_i$ share the same ancestors, the geometric distance between these points is affected by only their \textit{loop}s, as shown in \eqref{eq:d_1}. 
Thus, the geometric distance between points with the same ancestors in \eqref{eq:d_1} can decreases compared to that between points with different ancestors in \eqref{eq:d_2} as  $\norm{S_{A_l(p_s)}^l - S_{A_l(p_i)}^l}^2=0$.

\begin{figure*}[t]
  \centering
  \includegraphics[width=\linewidth]{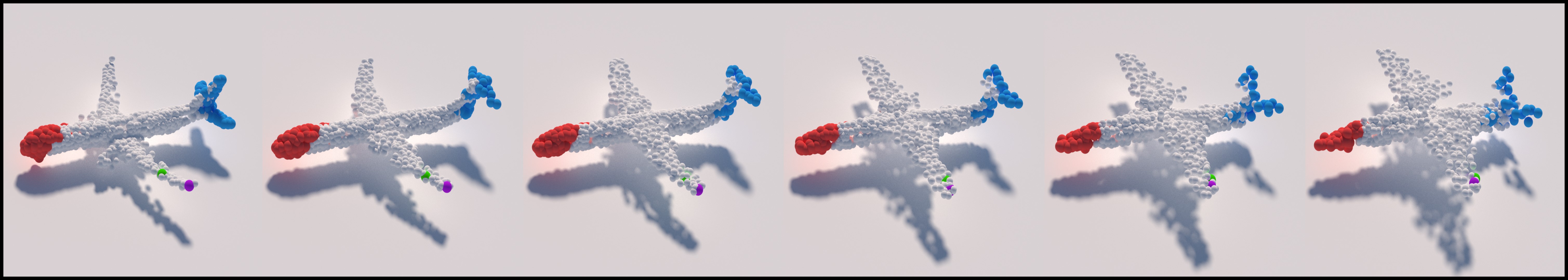}
  \caption{ \textbf{Semantic part generation and interpolation results of our tree-GAN.} Red and blue point clouds are generated from different ancestors in the tree, which form geometrically different families of points.
  The leftmost and rightmost point clouds of the airplanes were generated from different noise inputs. 
  The middle airplanes were obtained by interpolating between the leftmost and rightmost point clouds based on latent space representations.
}
  \label{fig:part}
\end{figure*}

Based on these two properties, our tree-GAN can generate semantic parts of objects, as shown in Fig.\ref{fig:part}, where points with the same ancestors are assumed to belong to the same parts of objects.   
We will explore this part generation problem for the proposed tree-GAN in Section~\ref{sec:part}.

%%%%%%%%%%%%%%%%%%%%%%%%%%%%%%%%%%%%%%%%%%%%%%%%%%%%%%%%%%%%%%%%%%%%%%%%%%%%%%%%%%%%%%%%%%
%%%%%%%%%%%%%%%%%%%%%%%%%%%%%%%%%%%%%%%%%%%%%%%%%%%%%%%%%%%%%%%%%%%%%%%%%%%%%%%%%%%%%%%%%%
\section{Fr\'echet Point Cloud Distance} \label{sec:FPD}
For quantitative comparisons between GANs, we require evaluation metrics that can accurately measure the quality of the 3D point clouds generated by GANs.
In the case of 2D data generation problems, FID~\cite{fid} is the most common metric.
FID adopts pre-trained inception V3 models~\cite{InceptionV3} to utilize their feature spaces for evaluation.
Although the conventional metrics proposed by Achlioptas \emph{et al.}~\cite{r-gan} can be used to evaluate the quality of generated points by directly measuring matching distances between real and generated point clouds, they can be considered as sub-optimal metrics because the goal of a GAN is not to generate the similar samples (\emph{e.g.}, MMD or CD) but to generate synthetic probability measures that are as close as possible to real probability measures. 
This perspective has been explored in unsupervised 2D image generation tasks using GANs~\cite{IS, fid}.
Therefore, we propose a novel evaluation metric for generated 3D point clouds called FPD.

Similar to FID, the proposed FPD calculates the 2-Wasserstein distance between real and fake Gaussian measures in the feature spaces extracted by PointNet~\cite{Qi17} as follows:
\begin{equation} \label{eq:fpd}
   \text{FPD}(\mathbb{P}, \mathbb{Q}) = \norm{\mathbf{m}_{\mathbb{P}} - \mathbf{m}_{\mathbb{Q}}}_2^2 + \text{Tr}( \Sigma_{\mathbb{\mathbb{P}}} + \Sigma_{\mathbb{\mathbb{Q}}} - 2(\Sigma_{\mathbb{\mathbb{P}}}\Sigma_{\mathbb{\mathbb{Q}}})^{\frac{1}{2}} ),
\end{equation}
where $\mathbf{m}_{\mathbb{P}}$ and $\Sigma_{\mathbb{P}}$ are the mean vector and covariance matrix of the points calculated from real point clouds $\{ x \}$, respectively, and $\mathbf{m}_{\mathbb{Q}}, \Sigma_{\mathbb{Q}}$ are the mean vector and covariance matrix calculated from generated point clouds $\{ x' \}$, respectively, where $x \sim \mathbb{P}$ and $x'=G(z) \sim \mathbb{Q}$.
In \eqref{eq:fpd}, $\text{Tr}(A)$ is the sum of the elements along the main diagonal of matrix $A$.
In this paper, for evaluation purposes, we use both conventional evaluation metrics~\cite{r-gan} and the proposed FPD.

%%%%%%%%%%%%%%%%%%%%%%%%%%%%%%%%%%%%%%%%%%%%%%%%%%%%%%%%%%%%%%%%%%%%%%%%%%%%%%%%%%%%%%%%%%
%%%%%%%%%%%%%%%%%%%%%%%%%%%%%%%%%%%%%%%%%%%%%%%%%%%%%%%%%%%%%%%%%%%%%%%%%%%%%%%%%%%%%%%%%%
\section{Experimental Results} \label{sec:exp}

\noindent \textbf{Implementation details: }
We used the Adam optimizer for both the generator and discriminator networks with a learning rate of $\alpha = 10^{-4}$ and other coefficients of $\beta_1 = 0$ and $\beta_2 = 0.99$. 
In generator, we used LeakyReLU as a nonlinearity function without batch normalization.
The network architecture of discriminator was the same as that in $r$-GAN~\cite{r-gan}. 
The gradient penalty coefficient was set to $10$ and the discriminator was updated five times per iteration, while the generator was updated one time per iteration.
As shown in Fig.\ref{fig:pipeline}, a latent vector $z \in \mathbb{R}^{96}$ was sampled from a normal distribution $\mathcal{N}(0,I)$ ) to act as an input.
Seven layers ($L=7$) were used for the TreeGCN.
The loop term of the TreeGCN in \eqref{eq:loop} had $K=10$ supports.
The total number of points in the final layer was set to $n=2048$.

\noindent \textbf{Comparison: }
There are only two conventional GANs for 3D point cloud generation: r-GAN~\cite{r-gan} and the GAN proposed by Valsesia \emph{et al.}~\cite{Localized-GCN}.
Thus, the proposed tree-GAN was compared to these two GANs. 
While the conventional GANs in~\cite{Localized-GCN, r-gan} train separate networks for each class, our tree-GAN trains only a single network for multiple classes of objects. 

\noindent \textbf{Evaluation metrics: }
We evaluated the \textbf{tree-GAN} using ShapeNet\footnote{\url{https://www.shapenet.org/}}, which is a large-scale dataset of 3D shapes, containing $16$ object classes. 
Evaluations were conducted in terms of the proposed FPD (Section~\ref{sec:FPD}) and the metrics used by Achlioptas et al~\cite{r-gan}.
As a reference model for FPD, we used the classification module of PointNet~\cite{Qi17} because it can handle partial inputs of objects.
This property is suitable for FPD because generated point clouds gradually form shapes, meaning point clouds can be partially complete during training.
For the implementation of FPD, we first trained a classification module for $40$ epochs to attain an accuracy of $98\%$ for classification tasks. 
We then extracted a $1808$-dimensional feature vector from the output of the dense layers to calculate the mean and covariance in \eqref{eq:fpd}.  
 
\begin{table*}[t]
\caption{\textbf{Quantitative comparison in terms of the metrics used by Achlioptas \emph{et al.}~\cite{r-gan}.} Red and blue values denote the best and the second-best results, respectively. Because the original paper by Valsesia \emph{et al.}~\cite{Localized-GCN} only presented point cloud results for chair and airplane classes, our tree-GAN was compared to \cite{Localized-GCN} based on these classes. 
However, we additionally evaluated the proposed tree-GAN quantitatively for all $16$ classes, as shown below. For networks with $\star$, we used results reported in~\cite{Localized-GCN}.
Higher COV-CD and COV-EMD, and lower JSD, MMD-CD, and MMD-EMD indicate better methods.}
\label{table:Comparison1}
\begin{center}
 \vspace{-3mm}
\setlength{\tabcolsep}{7pt}
        \begin{tabular}{ccccccc}
    \textbf{Class} & \textbf{Model} & \textbf{JSD $\downarrow$} & \textbf{MMD-CD $\downarrow$} & \textbf{MMD-EMD $\downarrow$} & \textbf{COV-CD $\uparrow$} & \textbf{COV-EMD $\uparrow$} \\ \hline
    
 & \textbf{r-GAN (dense)}$^{\star}$ 
 & 0.238 
 & {\textcolor{blue}{\textbf{0.0029}}} 
 & 0.136 
 & {\textcolor{blue}{\textbf{33}}} 
 & 13 \\
 
 & \textbf{r-GAN (conv)}$^{\star}$ 
 & 0.517 
 & 0.0030 
 & 0.223 
 & 23 
 & 4 \\
 
 & \textbf{Valsesia \emph{et al.} (no up.)}$^{\star}$ 
 & {\textcolor{blue}{\textbf{0.119}}} 
 & 0.0033 
 & 0.104 
 & 26 
 & 20 \\
 
 & \textbf{Valsesia \emph{et al.} (up.)}$^{\star}$ 
 & {\textcolor{red}{\textbf{0.100}}} 
 & {\textcolor{blue}{\textbf{0.0029}}}
 & {\textcolor{red}{\textbf{0.097}}} 
 & 30 
 & {\textcolor{blue}{\textbf{26}}} \\
 
\multirow{-5}{*}{\textbf{Chair}} 
 & \textbf{tree-GAN (Ours)} 
 & {\textcolor{blue}{\textbf{0.119}}} 
 & {\textcolor{red}{\textbf{0.0016}}} 
 & {\textcolor{blue}{\textbf{0.101}}} 
 & {\textcolor{red}{ \textbf{58}}} 
 & {\textcolor{red}{\textbf{30}}} \\ \hline
 
 & \textbf{r-GAN (dense)}$^{\star}$ 
 & 0.182 
 & 0.0009 
 & 0.094 
 & {\textcolor{blue}{\textbf{31}}} 
 & 9 \\
 
 & \textbf{r-GAN (conv)}$^{\star}$ 
 & 0.350 
 & {\textcolor{blue}{\textbf{0.0008}}}
 & 0.101 & 26 & 7 \\
 
 & \textbf{Valsesia \emph{et al.} (no up.)}$^{\star}$ 
 & 0.164 
 & 010010 
 & 0.102 
 & 24 
 & 13 \\
 
 & \textbf{Valsesia \emph{et al.} (up.)}$^{\star}$ 
 & {\textcolor{red}{\textbf{0.083}}} 
 & {\textcolor{blue}{\textbf{0.0008}}} 
 & {\textcolor{blue}{\textbf{0.071}}} 
 & {\textcolor{blue}{\textbf{31}}} 
 & {\textcolor{blue}{\textbf{14}}} \\
 
\multirow{-5}{*}{\textbf{Airplane}}
 & \textbf{tree-GAN (Ours)} 
 & {\textcolor{blue}{\textbf{0.097}}} 
 & {\textcolor{red}{\textbf{0.0004}}} 
 & {\textcolor{red}{\textbf{0.068}}} 
 & {\textcolor{red}{ \textbf{61}}} 
 & {\textcolor{red}{\textbf{20}}} \\ \hline

\multirow{-0.4}{*}{\textbf{All ($16$ classes)}}
 & \textbf{r-GAN (dense)} & 0.171 & 0.0021 & 0.155 & 58 & 29 \\
 & \textbf{tree-GAN (Ours)}
 & {\textcolor{red}{\textbf{0.105}}}
 & {\textcolor{red}{\textbf{0.0018}}}
 & {\textcolor{red}{\textbf{0.107}}}
 & {\textcolor{red}{\textbf{66}}}
 & {\textcolor{red}{\textbf{39}}} \\ \hline
\end{tabular}
%\vspace{-5mm}
\end{center}
\end{table*}

\begin{table}[t]
\caption{\textbf{Quantitative comparison in terms of the proposed FPD.} The FPD for the real point clouds was almost nearly zero.
This value can serve as the lower bound for the generated point clouds. Note that we could not evaluate the GAN proposed by Valsesia \emph{et al.}~\cite{Localized-GCN} in terms of FPD because the source code was not available. Better methods have smaller values of FPD. }
\label{table:Comparison2}
 \begin{center}
 \vspace{-5mm}
 +
\setlength{\tabcolsep}{10pt}
        \begin{tabular}{cccc}
    \textbf{Class} & \textbf{Model} & \textbf{FPD $\downarrow$} 
    \\ \hline
 & \textbf{r-GAN} & 1.860 & \\
\multirow{-2}{*}{\textbf{Chair}} & \textbf{tree-GAN (Ours)} & {\textcolor{red}{\textbf{0.809}}} \\ \hline
 & \textbf{r-GAN} & 1.016 \\
\multirow{-2}{*}{\textbf{Airplane}} & \textbf{tree-GAN (Ours)} & {\textcolor{red}{\textbf{0.439}}} \\ \hline
 & \textbf{r-GAN} & 4.726 \\
\multirow{-1}{*}{\textbf{All} ($16$ classes)} & \textbf{tree-GAN (Ours)} & {\textcolor{red}{\textbf{3.600}}} \\
& \textbf{Real (Low bound)} & \textbf{0} \\\hline
        \end{tabular}
        %\vspace{-5mm}
 \end{center}
\end{table}

%%%%%%%%%%%%%%%%%%%%%%%%%%%%%%%%%%%%%%%%%%%%%%%%%%%%%%%%%%%%%%%%%%%%%%%%%%%%%%%%%%%%%%%%%%
\subsection{Ablation Study} \label{sec:part}
We analyze the proposed tree-GAN and examine its useful properties, namely unsupervised semantic part generation, latent space representation via interpolation, and branching.  

\noindent \textbf{Unsupervised semantic part generation: }
Our tree-GAN can generate point clouds for different semantic parts, even with no prior knowledge regarding those parts during training. 
The tree-GAN can perform this semantic generation owing to its tree-structured graph convolution, which is a unique characteristic among GAN-based 3D point cloud methods.
As stated in Proposition~\ref{prop:1}, the geometric distance between points is determined by their ancestors in the tree. 
Different ancestors imply geometrically different families of points. 
Therefore, by selecting different ancestors, our tree-GAN can generate semantically different parts of point clouds.
Note that these geometric families of points are consistent between different latent code inputs. 
For example, let $z_1,z_2 \sim \mathcal{N}(0, I)$ be sampled latent codes. 
Let $G(z_1)=[p_1~p_2~\dots~p_{2048}]$ and $G(z_2)=[q_1~q_2~\dots~q_{2048}]$ be their corresponding generated point clouds. 
Let $J$ be a certain subset of $2048$ point indices. 
Then, $G^J(z_1)=[p_j]_{j \in J}$ denotes the subset of $G(z_1)$ from the indices $J$.
As shown in Fig.~\ref{fig:part}, if we select the same subsets of indices of point clouds, it results in the same semantic parts, even though the latent code inputs are different. 
For example, all red points indexed by $J_h$ (\emph{e.g.}, $G^{J_h}(z_1)$ and $G^{J_h}(z_2)$) represent the cockpits of airplanes, while all blue points indexed by $J_t$ (\emph{e.g.}, $G^{J_t}(z_1)$ and $G^{J_t}(z_2)$) represent the tails of airplanes.

From this ablation study, we can verify that the differences between ancestors determine the semantic differences between points and that two points with the same ancestor (\emph{e.g.}, green and purple points in the left wings in Fig.~\ref{fig:part}) maintain their relative distances for different latent codes. 

\noindent \textbf{Interpolation: }
We interpolated 3D point clouds by setting the input latent code to $z_\alpha = (1-\alpha)z_1 + \alpha z_2$ based on six alphas $\alpha=[\alpha_1, \dots, \alpha_6]$. 
The leftmost and rightmost point clouds of the airplanes in Fig.\ref{fig:part} were generated by $G(z_1)$ and $G(z_2)$, respectively. 
Our tree-GAN can also generate realistic interpolations between two point clouds.

\noindent \textbf{Branching strategy: } We conducted the experiments to show that the convergence dynamics of the proposed metric is not sensitive to different branching strategies.
Like other experiments, the total number of the generated points are $2048$ but different branching degrees were set (\emph{e.g.},
$\{d_l\}^7_1 = \{1,2,2,2,2,2,64 \}$, $ \{1,2,4,16,4,2,2 \}$, $ \{1,32,4,2,2,2,2 \}$). 
Please refer to convergence graphs in supplementary materials. 

%%%%%%%%%%%%%%%%%%%%%%%%%%%%%%%%%%%%%%%%%%%%%%%%%%%%%%%%%%%%%%%%%%%%%%%%%%%%%%%%%%%%%%%%%%
\subsection{Comparisons with Other GANs}
The proposed tree-GAN was quantitatively and qualitatively compared to other state-of-the-art GANs for point cloud generation, in terms of both accuracy and computational efficiency. 
Supplementary materials contain more results and comparisons for 3D point clouds generation.

\noindent \textbf{Comparisons: }
Tables~\ref{table:Comparison1} and \ref{table:Comparison2} contain quantitative comparisons in terms of the metrics used by Achlioptas \emph{et al.}~\cite{r-gan} (\emph{i.e.}, JSD, MMD-CD, MMD-EMD, COV-CD, and COV-EMD) and the proposed FPD, respectively.
The proposed tree-GAN consistently outperforms other GANs at a large margin in terms of all metrics, demonstrating the effectiveness of the proposed treeGCN.

\begin{figure*}[t]
  \begin{center}
    \includegraphics[width=\linewidth]{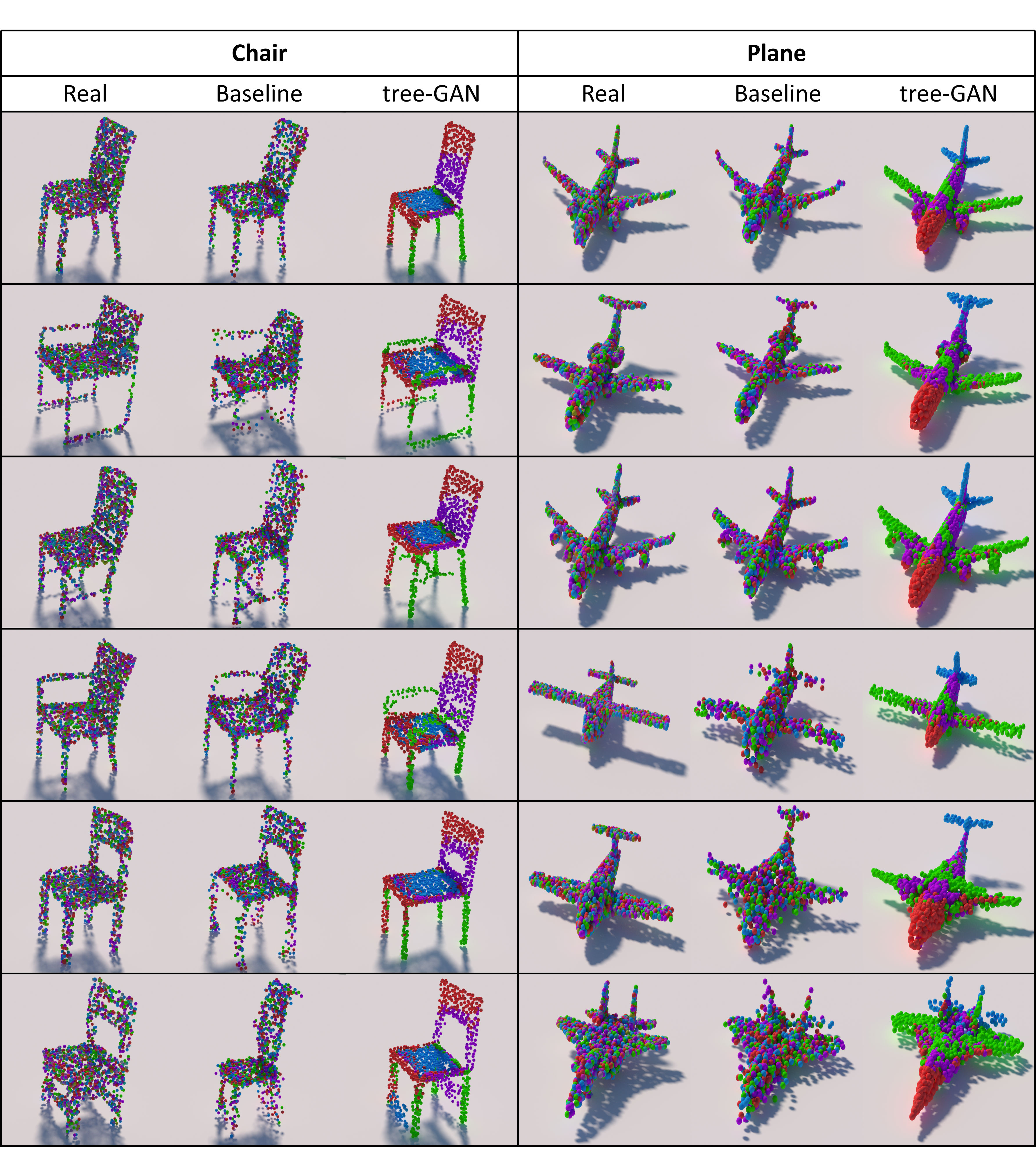}
  \end{center}
  \vspace{-5mm}
  \caption{ \textbf{Unsupervised 3D point cloud generation results of baseline (\emph{i.e.} r-GAN~\cite{r-gan}) and our tree-GAN.} The proposed tree-GAN generates more accurate and detailed point clouds of objects as comparison with r-GAN, and produces point clouds for each part of the objects even with no prior knowledge on that part. 
  The point clouds generated by the tree-GAN can represent a variety of geometrical typologies for each class. The first, second, and third columns show point clouds of ground truth, baseline, and tree-GAN, respectively.}
  \label{fig:all_class}
\end{figure*}

For qualitative comparisons, we equally divided the entire index set into four subsets and painted the points in each subset with the same color. 
Although the real 3D point clouds were unordered as shown in Figs.\ref{fig:teaser} and \ref{fig:all_class}, our tree-GAN successfully generated 3D point clouds with intuitive semantic meaning without any prior knowledge, whereas r-GAN failed to generate semantically ordered point clouds. 
Additionally, our tree-GAN could generate detailed and complex parts of objects, whereas r-GAN generated more dispersed point distributions. 
Fig.\ref{fig:all_class} presents qualitative results of our tree-GAN.
The tree-GAN generated realistic point clouds for multi-object categories and produced very diverse typologies of point clouds for each class. 
 
\noindent \textbf{Computational cost: }
In methods using \textit{static links} for graph convolution, adjacency matrices are typically used for the convolution of vertices. 
Although these methods are known to produce good results for graph data, prior knowledge regarding connectivity is required.
In other methods that use \textit{dynamic links} for graph convolution, adjacency matrices must be constructed from vertices to derive connectivity information for every convolution layer instead of using prior knowledge. 
For example, let $L,B,V_l$ denote the number of layers, batch size, and induced vertex size of an output graph at the $l$-th layer, respectively.  
The methods described above require additional computations to utilize connectivity information.  
These computations require time and memory resources on the order of $\sum_{l=1}^L B \times V_l \times V_l$. 
However, our TreeGCN does not require any prior connectivity information like \textit{static link} methods and does not require additional computation like \textit{dynamic link} methods. 
Therefore, our network can use time and memory resources much more efficiently and requires less resources on the order of $\sum_{l=1}^L B \times V_l$.

~

~

~

~

~

%%%%%%%%%%%%%%%%%%%%%%%%%%%%%%%%%%%%%%%%%%%%%%%%%%%%%%%%%%%%%%%%%%%%%%%%%%%%%%%%%%%%%%%%%%
%%%%%%%%%%%%%%%%%%%%%%%%%%%%%%%%%%%%%%%%%%%%%%%%%%%%%%%%%%%%%%%%%%%%%%%%%%%%%%%%%%%%%%%%%%
\section{Conclusion} \label{sec:con}
In this paper, we proposed a generative adversarial network called the tree-GAN that can generate 3D point clouds in an unsupervised manner. 
The proposed generator for tree-GAN, which is called tree-GCN, preforms graph convolutions based on tree structures. 
The tree-GCN utilizes ancestor information from a tree and employs multiple supports to represent 3D point clouds. 
Thus, the proposed tree-GAN outperforms other GAN based point cloud generation methods in terms of accuracy and computational efficiency. 
Through various experiments, we demonstrated that the tree-GAN can generate semantic parts of objects without any prior knowledge and can represent 3D point clouds in latent spaces via interpolation.

{\small
\bibliographystyle{ieee}
\bibliography{egbib}
}
 
\end{document}